\documentclass[conference]{IEEEtran}
\IEEEoverridecommandlockouts
\usepackage{cite}
\usepackage{amsmath,amssymb,amsfonts}
\usepackage{graphicx}
\usepackage{textcomp}
\usepackage{xcolor}
\usepackage{algpseudocode}
\usepackage{amsmath}
\usepackage{amsfonts}
\usepackage{booktabs}
\usepackage[shortlabels]{enumitem}
\usepackage{amssymb} 
\usepackage[linesnumbered,ruled,vlined]{algorithm2e}

\def\BibTeX{{\rm B\kern-.05em{\sc i\kern-.025em b}\kern-.08em
    T\kern-.1667em\lower.7ex\hbox{E}\kern-.125emX}}
\begin{document}

\title{Evolutionary Neural Architecture Search for 3D Point Cloud Analysis}


\author{\IEEEauthorblockN{Yisheng Yang\IEEEauthorrefmark{2}, Guodong Du\IEEEauthorrefmark{3}, Chean Khim Toa\IEEEauthorrefmark{2}, Ho-Kin Tang\IEEEauthorrefmark{3}, Sim Kuan Goh\IEEEauthorrefmark{4}}
\IEEEauthorblockA{\IEEEauthorrefmark{2}School of Computing and Data Science, Xiamen University Malaysia, Sepang, Malaysia.}
\IEEEauthorblockA{\IEEEauthorrefmark{3}School of Science, Harbin Institute of Technology~(Shenzhen), China.}
\IEEEauthorblockA{\IEEEauthorrefmark{4}School of Electrical Engineering and Artificial Intelligence,
Xiamen University, Malaysia. \\}
\thanks{This work was supported in part by Shenzhen College Stability Support Plan (GXWD20231128103232001), Department of Science and Technology of Guangdong (2024A1515011540), Shenzhen Start-Up Research Funds~(HA11409065), National Natural Science Foundation of China (12204130), the Ministry of Higher Education Malaysia through the Fundamental Research Grant Scheme (FRGS/1/2023/ICT02/XMU/02/1), and Xiamen University Malaysia Research Fund (XMUMRF/2022-C10/IECE/0039 and XMUMRF/2024-C13/IECE/0049). \ 
Corresponding: denghaojian@hit.edu.cn, simkuan.goh@xmu.edu.my}}

\maketitle

\begin{abstract}
Neural architecture search (NAS) automates neural network design by using optimization algorithms to navigate architecture spaces, reducing the burden of manual architecture design. While NAS has achieved success, applying it to emerging domains, such as analyzing unstructured 3D point clouds, remains underexplored due to the data lying in non-Euclidean spaces, unlike images. This paper presents Success-History-based Self-adaptive Differential Evolution with a Joint Point Interaction Dimension Search (SHSADE-PIDS), an evolutionary NAS framework that encodes discrete deep neural network architectures to continuous spaces and performs searches in the continuous spaces for efficient point cloud neural architectures. Comprehensive experiments on challenging 3D segmentation and classification benchmarks demonstrate SHSADE-PIDS's capabilities. It discovered highly efficient architectures with higher accuracy, significantly advancing prior NAS techniques. For segmentation on SemanticKITTI, SHSADE-PIDS attained 64.51\% mean IoU using only 0.55M parameters and 4.5GMACs, reducing overhead by over 22-26X versus other top methods. For ModelNet40 classification, it achieved 93.4\% accuracy with just 1.31M parameters, surpassing larger models. SHSADE-PIDS provided valuable insights into bridging evolutionary algorithms with neural architecture optimization, particularly for emerging frontiers like point cloud learning.

\end{abstract}


\section{Introduction}
Deep neural networks have achieved remarkable success across diverse applications, largely attributed to their multi-layered architecture enabling hierarchical feature learning \cite{du2024knowledge,jiang2023enhancing,r3,goh2018spatio,goh2016multiway}. However, the design of network architectures relies heavily on expert experience through extensive experiments. This manual process is computationally expensive and time-consuming due to the need to train and assess a vast number of configurations. To address this challenge, neural architecture search (NAS) has emerged as a promising technique to automate neural network design. The key idea is to search for an optimal architecture within a predefined search space by using an optimization algorithm that searches for candidate architectures \cite{r1}. NAS methods have discovered novel architectures surpassing hand-designed networks, including NASNet for image classification.

Point clouds captured by LiDAR and depth sensors are increasingly ubiquitous, but their unstructured nature poses difficulties for standard deep learning methods designed for grid data. Hence, applying NAS to emerging domains like point cloud processing remains an open challenge. While specialized point-based networks have been proposed, their manual architecture engineering is onerous. This provides strong motivation for investigating NAS tailored to automate point cloud network design.

Evolutionary algorithms provide an alternative and efficient search strategy by simulating a population of neural architectures that evolve for survival and breeding while solving the learning tasks. Popular methods include genetic algorithms, evolutionary strategies, genetic programming, differential evolution, particle swarm optimization, and ant colony optimization. In particular, differential evolution (DE) through adaptive control has shown promise in balancing exploration and exploitation \cite{qin2005self,tang2021yi}, leveraging additional historical evaluations to guide the search in large search spaces.

In this work, we (i) present a NAS framework driven by a proposed adaptive DE algorithm, Success-History-based Self-adaptive Differential Evolution (SHSADE) with (ii) a Joint Point Interaction Dimension Search (PIDS)~\cite{zhang2023pids} to (iii) fit geometric dispositions and density distributions in varying 3D point clouds. The hybridized SHSADE-PIDS enables optimized point cloud architectures specialized for tasks like segmentation and classification by encoding discrete deep neural network architectures to continuous spaces and performing searches in the continuous spaces. Comprehensive experiments demonstrate SHSADE-PIDS's optimizer capabilities and advantages in accelerating NAS. The results provide insights into effectively applying adaptive evolution strategies for architecture search, particularly in novel frontiers involving unstructured 3D data.

\begin{figure*}
    \centering
    \includegraphics[width=1\linewidth]{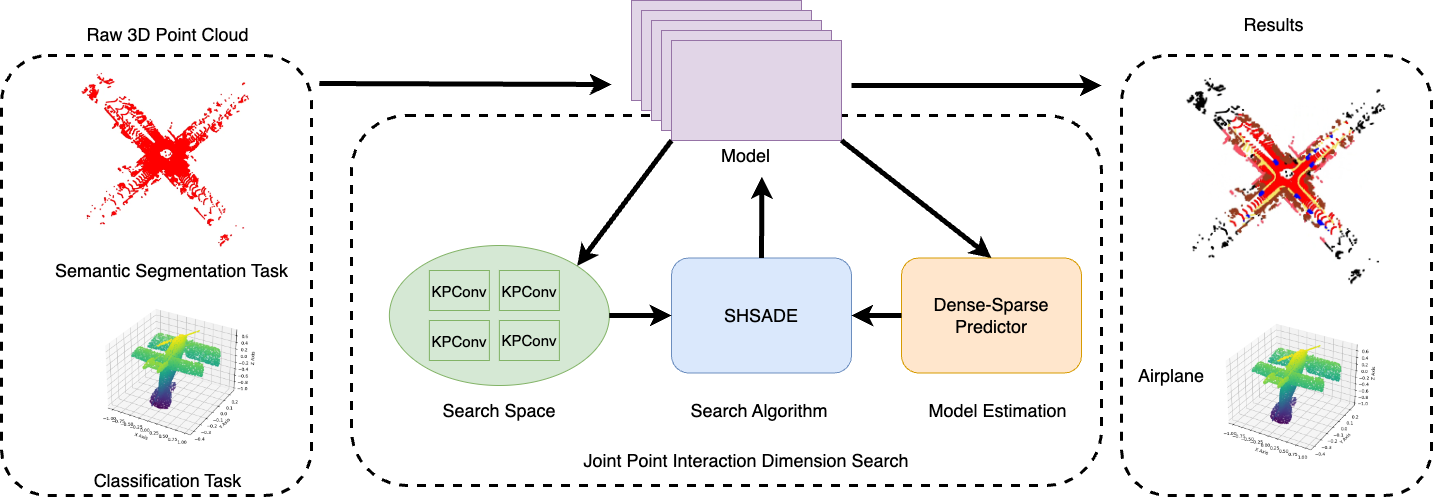}
    \caption{Proposed framework of Neural Architecture Search for 3D Point Cloud Learning Tasks via SHSADE with the encoding of discrete architectures in continuous space,  Joint Point Interaction Dimension Search (PIDS), and dense-sparse predictor.}
    \label{fig:enter-label}
\end{figure*}

\section{Related works}

\subsection{Neural Architecture Search}
Neural architecture search (NAS) has emerged as an important field of automated machine learning, aiming to automate the design of optimal neural network architectures tailored to specific tasks and datasets \cite{real2019regularized}. 
The key idea behind NAS is to define a search space encompassing possible network architectures and utilize an optimization algorithm to actively navigate this space to find high-performing architectures for a given task. The search algorithm trains and evaluates candidate architectures within the space to guide the search toward optimal solutions based on performance metrics on a validation set. By effectively exploring the vast search space guided by such empirical performance feedback, NAS methods can discover novel architectures surpassing human-designed counterparts specialized for the dataset and application.

Evolutionary algorithms have emerged as a particularly promising approach to neural architecture search due to their population-based search mechanism, which is naturally suited for exploring the vast combinatorial spaces of neural architectures. Notable evolutionary NAS techniques include NeuroEvolution of Augmenting Topologies (NEAT) \cite{chen2006neuroevolution}, Genetic CNN \cite{xie2017genetic}, and more recently AmoebaNet \cite{shah2018amoebanet} which leverages evolution strategies for architecture search. Compared to reinforcement learning-based NAS methods, evolutionary approaches have been shown to discover highly competitive architectures with greater stability, lower computational requirements, and less hyperparameter tuning. As such, evolutionary algorithms provide an efficient and powerful search strategy to automate neural architecture optimization without prohibitive computational costs.

\subsection{Differential Evolution Algorithm}
Differential Evolution (DE) is an evolutionary optimization algorithm designed to search continuous solution spaces for global optima. First proposed by Storn and Price in 1995, DE operates on real-valued solution vectors that are evolved over generations through bio-inspired mutation, crossover, and selection operators. The key advantage of DE is its self-adaptive mutation scheme that automatically adjusts the step size and direction during optimization based on differences between randomly selected population members. This allows DE to effectively balance the exploration and exploitation of complex problems.





A key limitation is DE's sensitivity to fixed control parameters, scale factor F, and crossover rate CR, which require problem-specific tuning. Since its introduction, numerous enhancements to DE have been developed, including the Success-History based Adaptive Differential Evolution (SHADE) algorithm. SHADE \cite{tanabe2013success} enhances adaptation through historical memories of CR and F that record statistics of successful F and CR values over multiple generations. Weighted average adaptation has a higher influence on more successful solutions. By utilizing memories to guide reliable tuning, SHADE improves optimization speed, accuracy, and robustness compared to fixed parameter DE.

Self-Adaptive DE (SADE) addresses this by adapting F and CR during execution based on their relative success over a learning period \cite{qin2005differential}. However, SADE's adaptation scheme has limitations regarding sensitivity to randomness and differentiation between moderately and highly successful values.

\subsection{Deep learning methods for point cloud processing}
Point clouds have become a prevalent 3D data representation, enabling detailed digitization of object shapes and environments through dense geometric sampling. However, point clouds present unique challenges for applying deep neural networks designed for grid-structured data like images. Point clouds are unordered, unstructured, and lack inherent topology. Pioneering deep learning approaches that directly process raw point clouds while respecting their permutation invariant and unstructured nature have emerged in recent years.

Early deep learning methods converted point clouds to intermediary representations like voxels or multi-view projections to leverage established 3D and 2D convolutional architectures. However, these conversions introduce quantization errors and lose fine geometric details. PointNet \cite{qi2017pointnet} pioneered direct feature learning on raw point sets using shared multilayer perceptrons (MLPs) and global max pooling for permutation invariance.

Various specialized point convolution operators have also been introduced to improve local feature learning, such as continuous convolutions in Kernel Point Convolution (KPConv) \cite{thomas2019kpconv}. Graph neural networks are gaining interest by inherently modeling connectivity in points through techniques like graph pooling and attention \cite{wang2019dynamic2}. Overall, point-based deep networks now achieve results by directly operating on unstructured point clouds without conversions.

KPConv is a pioneering point convolution operator using a parameterized kernel with learnable deformations to achieve geometry-adapted feature extraction. By representing kernels as sets of points with influence defined by distance correlations, KPConv enables flexible and effective point cloud feature learning. Networks using KPConv layers have achieved excellent performance on shape classification, part segmentation, and scene analysis benchmarks. The intuitiveness of learning convolutions directly on raw point sets and the deformable nature of KPConv have made it an indispensable building block for many point-based deep-learning architectures.

\subsection{Joint Point Interaction Dimension Search}
Point clouds provide detailed 3D geometric representations captured by sensors like LiDARs. However, their unstructured nature poses challenges for standard deep-learning techniques designed for grid data~\cite{9931941,goh2021tunnel}. Point-based deep neural networks have been proposed to process raw point clouds while retaining geometric details directly. However, manually designing optimal architectures for this domain remains difficult.

Recent works have proposed various strategies to design point operators, which are the basic building blocks of hierarchical 3D models, to extract features from point clouds. These strategies include point convolutions, graph neural networks, attention mechanisms, and kernel point convolutions \cite{guo2021pct}. However, existing approaches have limitations. First, they manually design a single type of point interaction and reuse it within all point operators of a 3D model, which may limit performance due to varying geometric/density distributions of points. Second, they optimize point interactions and dimensions separately, missing the opportunity to find better combinations on the two axes. To address these issues, Zhang et al. introduced PIDS (Joint Point Interaction Dimension Search)\cite{zhang2023pids}, which jointly explores point interactions and dimensions in a large search space to craft 3D models that balance performance and efficiency for point clouds.

\textbf{Search Space:} PIDS is based on kernel point convolutions and introduces high-order point interactions to fit geometric dispositions and density distributions in varying 3D points. It also incorporates width, depth, and expansion ratio search components for efficient point dimensions. The joint search space over these heterogeneous axes enables the discovery of networks customized to geometric and density variations.


\textbf{Predictor Model:} To enable efficient search, PIDS uses a Dense-Sparse Predictor to accurately model the heterogeneous search space. The predictor learns separate dense and sparse embeddings to handle the continuous and categorical components. A dot product enables cross-feature communication. This unified encoding improves the prediction of performance in the joint PIDS search space compared to standard predictors.

\textbf{Bi-objective Neural Architecture Search:}
To achieve both high accuracy and efficiency, PIDS formulates neural architecture search as a bi-objective optimization problem~\cite{goh2016decompositional}, maximizing predictive accuracy while minimizing computational cost (FLOPs). Predictor models estimate accuracy and FLOPs to avoid expensive training. The bi-objective function balances accuracy versus efficiency to find optimal architectures with different trade-offs. By jointly exploring point interactions and dimensions in a large search space using an optimized predictor model and bi-objective search, PIDS can discover high-performing and efficient models specialized for processing 3D point cloud data.

\section{Methods}

\begin{figure*}[ht]
\centering
\includegraphics[width=0.95\textwidth]{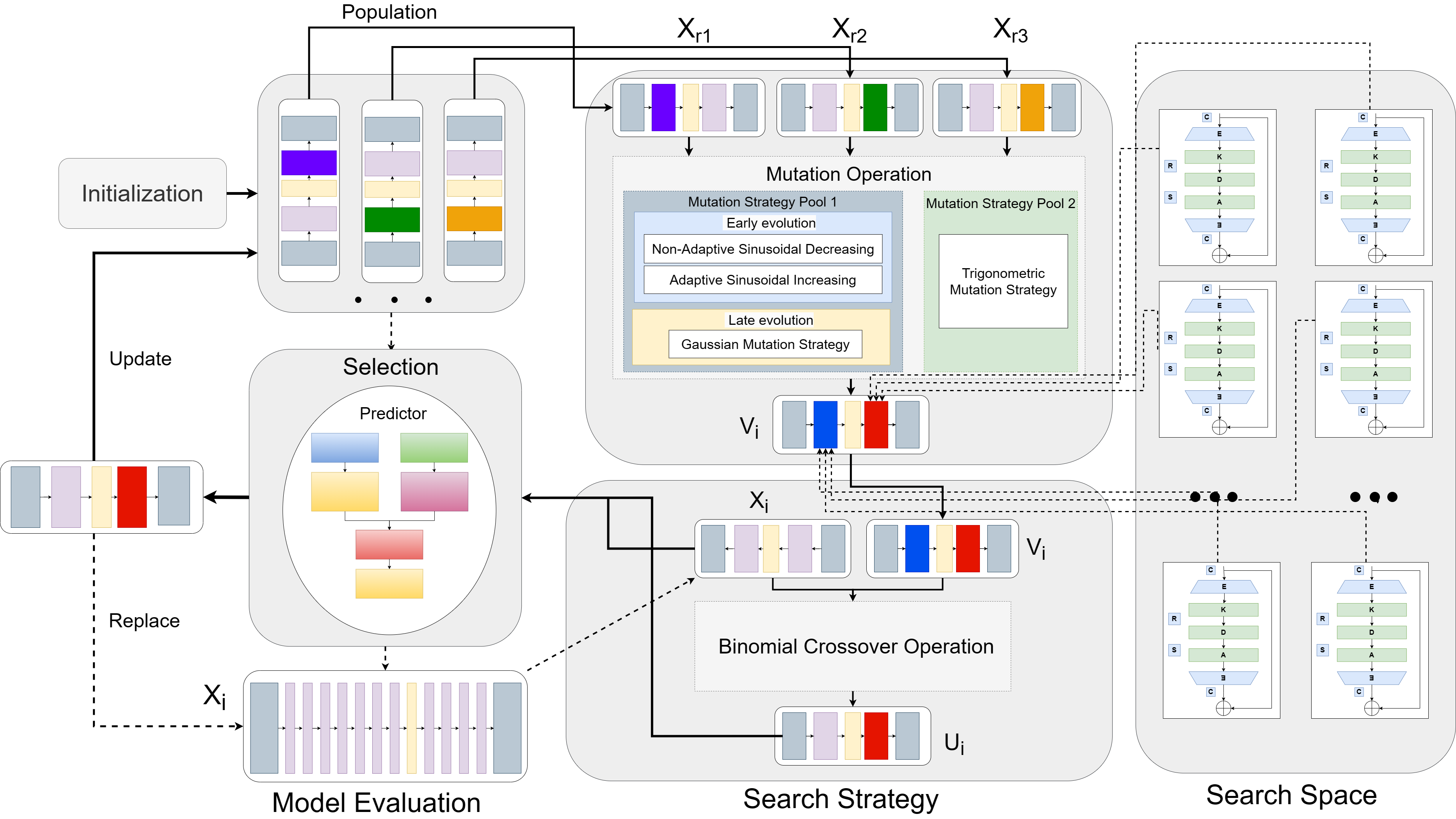}
\caption{Pipeline of the proposed SHSADE-PIDS for 3D point cloud, which mainly comprises the design of model evaluation, search strategy, and search space.}
\label{fig:shade-pids}
\end{figure*}

This section proposes a novel neural architecture search algorithm bridging discrete and continuous optimization, SHSADE-PIDS, which 
integrates techniques from SADE and SHADE, and describes its application in the 3D point cloud processing, illustrated in Fig.~\ref{fig:enter-label} and detailed in Fig.~\ref{fig:shade-pids}.

\subsection{SHSADE}\label{AA}
Differential Evolution (DE) is a popular evolutionary algorithm for real-parameter optimization. However, its performance depends significantly on properly setting control parameters like mutation factor (F) and crossover rate (CR). To overcome this limitation, this research proposes an enhanced DE variant called SHSADE, which adapts these parameters based on historical success.

SHSADE integrates complementary adaptive techniques from two major differential evolution variants - Self-Adaptive Differential Evolution (SADE) and Success-History based Adaptive Differential Evolution (SHADE). This hybridization aims to enhance the algorithm's optimization capabilities by leveraging the strengths of both approaches.

Specifically, from SADE, SHSADE adopts the core idea of self-adaptively tuning key control parameters like crossover rate (CR) and mutation factor (F) for each individual solution based on learning from historical memories. SADE records successful CR values over a predefined learning period and samples new CR values for each individual from a normal distribution with the mean set to the historically successful mean. This enables CR to be adapted to match the problem landscape. A similar mechanism is used to allow the mutation strength F to be tuned based on past successful values. Additionally, SADE maintains probabilities for selecting between multiple mutation strategies based on their relative success rates. This allows the algorithm to automatically choose the more effective strategies.

Complementarily, from SHADE, SHSADE inherits the use of memories to accumulate statistical summaries of successful control parameter values over multiple generations. Unlike SADE, which uses raw values, SHADE stores mean CR and F values from previous generations in historical memories of MCR and MF. New CR and F values are generated by randomly indexing into these memories and sampling from distributions around those stored means. The memories are then updated each generation by incorporating the new successful mean values using a learning rate. This more stable approach prevents the algorithm from overreacting to single generations where poor values succeed randomly.

Additionally, SHSADE integrates SHADE's current-to-pbest/1 mutation strategy along with novel trigonometric and sinusoidal strategies. The p-best selection helps balance greediness during the search.

\begin{enumerate}[(a)] 
\item \textbf{Historical Memories} Unlike DE, which uses fixed F and CR values, SHSADE maintains memories MCR and MF to store statistical summaries of successful parameter values from previous generations:

\begin{align*}
\mathbf{MCR} &= [\mathbf{MCR}_1, \mathbf{MCR}_2, \ldots, \mathbf{MCR}_H] \\
\mathbf{MF} &= [\mathbf{MF}_1, \mathbf{MF}_2, \ldots, \mathbf{MF}_H]
\end{align*}

\noindent where H is the memory size. Rather than raw values, these memories store the mean successful F and CR values from each of the past H generations (Eq. 1,2).

\begin{equation}
CR_i = \text{randn}(MCR_k, 0.1)
\end{equation}

\begin{equation}
F_i = \text{randc}(MF_k, 0.1)
\end{equation}

New F and CR values are generated by randomly indexing into these memories and sampling from distributions around the stored means (Eq. 3, 4). This allows exploiting knowledge from prior successful generations.
\begin{equation}
MCR_{k,G+1} = 
\begin{cases} 
\text{mean}(SCR) & \text{if } SCR \neq \emptyset \\
MCR_{k,G} & \text{otherwise}
\end{cases}
\end{equation}

\begin{equation}
MF_{k,G+1} = 
\begin{cases} 
\text{mean}(SF) & \text{if } SF \neq \emptyset \\
MF_{k,G} & \text{otherwise}
\end{cases}
\end{equation}

Sampling F and CR:

\begin{equation}
CR_i = \text{randn}(MCR_{r_i}, 0.1)
\end{equation}

\begin{equation}
F_i = \text{randc}(MF_{r_i}, 0.1)
\end{equation}

Where ri is a random index into the memories.
\item \textbf{Adaptive Parameter Control} After each generation, the memories MCR and MF are updated with the mean successful F and CR values using a learning rate c. This allows gradual adaptation of the control parameters to suit the problem landscape.
\item \textbf{Trigonometric Mutation Strategy} The trigonometric mutation strategy\cite{fan2003trigonometric}, hereafter referred to as Strategy 4, introduces an efficient approach in evolutionary algorithms by generating a new trial vector through a weighted averaging process involving three distinct members of the population, with consideration given to their relative fitness. This method leverages geometric relationships among the selected individuals, utilizing the concept of centroid and linear combinations in the objective function space.

Consider three distinct individuals $X_{r_1,G}$, $X_{r_2,G}$, and $X_{r_3,G}$ randomly selected from the current population, where $r1 \neq r2 \neq r3 \neq i$. The mutation for the $i$-th individual $X_{i, G}$ in generation $G+1$ is given by:


\begin{align*}
V_{i,G+1}
    & :=  \frac{X_{r_1,G} + X_{r_2,G} + X_{r_3,G}}{3} \\
    &  + (w_2 - w_1)(X_{r_1,G} - X_{r_2,G})  \\
    &  + (w_3 - w_2)(X_{r_2,G} - X_{r_3,G}) \\
    &  + (w_1 - w_3)(X_{r_3,G} - X_{r_1,G})
\end{align*}


where the weights $w_1$, $w_2$, and $w_3$ are defined as:
\begin{equation}
w_1 = \frac{|f(X_{r_1,G})|}{w^\prime}, \quad w_2 = \frac{|f(X_{r_2,G})|}{w^\prime}, \quad w_3 = \frac{|f(X_{r_3,G})|}{w^\prime}
\end{equation}
and
\begin{equation}
w^\prime = |f(X_{r_1,G})| + |f(X_{r_2,G})| + |f(X_{r_3,G})|
\end{equation}

The methodology incorporates a perturbation vector that is a linear combination of the triangle's legs, with weights corresponding to the objective function values at the vertices. This perturbation can be conceptualized as shifts by the triangle's center (the donor) along each leg, with varying step sizes.

The weight terms $(w_2-w_1)$, $(w_3-w_2)$, and $(w_1-w_3)$ serve a dual purpose. Firstly, they bias the movement along the triangle's legs from worse to better solutions based on the objective function. Secondly, they automatically scale the vector components in accordance with the differences in objective values among the individuals, thereby guiding the solution towards the most favorable of the three points. As such, the trigonometric mutation acts as a local search operator confined to the trigonometric region defined by the trio of individuals.

To illustrate this, consider a two-dimensional minimization problem with individuals $X_{r_1,G}$, $X_{r_2,G}$, and $X_{r_3,G}$. The boundaries of the trigonometric region can be determined by examining extreme cases of weight values, resulting in points $X_{t_1}^\prime$, $X_{t_2}^\prime$, and $X_{t_3}^\prime$. Consequently, any mutated individual will reside within this defined trigonometric region. The process, systematically moves the solution towards lower objective values, effectively exploiting the objective function information to guide the search towards promising areas.

The trigonometric mutation strategy's notable features include its simple geometric interpretation and the ability to automatically balance the differential vectors, which are key in guiding evolutionary algorithms towards efficient search paths in the solution space.

\begin{algorithm}[]
\caption{SHSADE}
\DontPrintSemicolon  
\tcp{Initialize population $\mathbf{P}$ with random solutions within bounds}
\tcp{Initialize memories: $M_F, M_{CR}, M_{\text{freq}}$}
\tcp{Set termination criteria: max generations $G_{\text{max}}$, convergence threshold, etc.}
\tcp{Initialize current generation $g = 1$}
\While{$g \leq G_{\text{max}}$ and not reached termination criteria}{
    \For{each individual $\mathbf{x}_i$ in population $\mathbf{P}$}{
        \tcp {Select mutation strategy $s$ based on strategy probabilities}
        \eIf{$s \neq \text{trigonometric}$}{
            \eIf{$g \leq \frac{G_{\text{max}}}{2}$}{
                \tcp{Update $F_i$ using sinusoidal or adaptive sinusoidal strategy}
            }{
                \tcp{ Update $F_i$ using Gaussian mutation: $F_i = \mathcal{N}(\mu_F, \sigma_F^2)$}
            }
            \tcp{ Update $CR_i$ using normal distribution: $CR_i = \mathcal{N}(\mu_{CR}, \sigma_{CR}^2)$}
        }{}
        \tcp{ Generate trial vector $\mathbf{u}_{i,g}$ using $s$, $F_i$, $CR_i$}
        \tcp{ Apply boundary check and regularization to $\mathbf{u}_{i,g}$}
        \If{fitness($\mathbf{u}_{i,g}$) $\leq$ fitness($\mathbf{x}_{i,g}$)}{
            \tcp{ Replace $\mathbf{x}_{i,g}$ with $\mathbf{u}_{i,g}$ in $\mathbf{P}$}
        }
    }
    \tcp{ Update memories based on success rates}
    $g = g + 1$\;
}
\Return Best solution found in $\mathbf{P}$
\end{algorithm}

\item \textbf{Mutation Operation in SHSADE Algorithm} The SHSADE algorithm incorporates a critical mutation operation that generates trial vectors by modifying existing population members. This operation is instrumental in exploring the solution space, injecting diversity, and helping the algorithm to escape local optima. It leverages a combination of mutation strategies, selected probabilistically based on their historical performance.

Among the employed strategies, the Sinusoidal Mutation strategy stands out\cite{awad2016ensemble}. These strategies adapt the scaling factor $F$ using sinusoidal functions to balance exploration and exploitation throughout the optimization process. Specifically, two sinusoidal approaches are mixed for the first half of the generations, $g_{s_1} \in [1, \frac{G_{\mathrm{max}}}{2}]$, to adapt $F_{i,g}$.

The Non-Adaptive Sinusoidal Decreasing Adjustment (Strategy 1) employs a decreasing sinusoidal formula to modify $F_{i,g_{s_1}}$ for each individual:
\begin{equation}
F_{i,g_{s_1}} = \frac{1}{2} \left( \sin(2\pi \cdot \mathrm{freq} \cdot g_{s_1} + \pi) \cdot \frac{G_{\mathrm{max}} - g_{s_1}}{G_{\mathrm{max}}} + 1 \right)
\end{equation}
Here, $\mathrm{freq}$ is a fixed frequency value, $g_{s_1}$ is the current generation, and $G_{\mathrm{max}}$ is the maximum number of iterations.

The Adaptive Sinusoidal Increasing Adjustment (Strategy 2) uses a different sinusoidal formula for an incremental adjustment of the scaling factor:
\begin{equation}
F_{i,g_{s_1}} = \frac{1}{2} \left( \sin(2\pi \cdot \mathrm{freq}_{i,g_{s_1}} \cdot g_{s_1}) \cdot \frac{g_{s_1}}{G_{\mathrm{max}}} + 1 \right)
\end{equation}
In this approach, $\mathrm{freq}_{i,g_{s_1}}$ is an adaptive frequency, adjusted each generation using a Cauchy distribution:
\begin{equation}
\mathrm{freq}_{i,g_{s_1}} = \mathrm{randc}(\mu\mathrm{freq}_{r_i,g_{s_1}}, 0.1)
\end{equation}
The parameter $\mu\mathrm{freq}_{r_i,g_{s_1}}$ for generation $g_{s_1}$ is selected from an external memory $M_{\mathrm{freq}}$, which stores average successful frequencies from previous generations in $S_{\mathrm{freq}}$. After each generation $g_{s_1}$, this parameter is updated based on the Lehmer mean at a randomly chosen index $r_i$.

These strategies exemplify the algorithm's approach to dynamically adapting mutation parameters, contributing to its effectiveness in complex optimization tasks.
\end{enumerate}





\begin{table*}[h]
\centering
\caption{mIOU on sequence 08 (validation split) of SemanticKITTI. Here, *: Results based on retrained on NVIDIA RTX4090}
\label{table:semantic_kitti}
\begin{tabular}{cccccc}
\hline
Architecture & Method & Params (M) & MACs(G)& Latency (ms)&mIOU(\%)\\ 
\hline
 Manual \\
 KPConv-rigid (Zhang et al., 2023) & Point-based & 14.8 & 60.9 & 221 (164 + 57) & 59.2 \\
 PIDS (second-order) (Zhang et al., 2023) & Point-based & 0.97 & 4.7 & 160 (103 + 57) & 60.1 \\
 SalsaNext (Cortinhal et al., 2020) & Projection-based & 6.7 & 62.8 & 71 & 59.0 \\
 MinkowskiNet (Choy et al., 2019) & Voxel-based & 5.5 & 28.5 & 294 & 58.9 \\
 NAS \\
 SPVNAS (Tang et al., 2020) & Voxel-based & 3.3 & 20.0 & 158 & 61.5 \\
 & & 7.0 & 34.7 & 175 & 63.5 \\
 PIDS (NAS) (Zhang et al., 2023) & Point-based & 0.57* & 4.4* & 125 (99 + 26)* & 62.4* \\
 PIDS (NAS 2X) (Zhang et al., 2023) & & 1.36* & 11.0* & 144 (103 + 41)* & 64.1* \\
 \textbf{SHSADE-PIDS(NAS, 1.25X)} & Point-based & \textbf{0.55} & \textbf{4.5} & \textbf{119 (99 + 23)} & \textbf{63.29} \\
 \textbf{SHSADE-PIDS (NAS, 2X)} & & \textbf{1.36} & \textbf{8.6} & \textbf{132 (103 + 29)} & \textbf{64.51} \\
\hline
\end{tabular}
\end{table*}

\begin{table*}[h]
\centering
\caption{mIOU on sequence 08 (validation split) of SemanticKITTI. Here, *: Results based on retrained on NVIDIA RTX4090}
\label{table:modelnet40}
\begin{tabular}{cccccc}
\hline
 Method & Params (M) & Latency (ms)& Overall Accuracy (\%)
 \\ 
\hline
 Manual \\
PointNet (Qi et al., 2017) & - & - & 89.2 \\
SO-Net (Li et al., 2018) & - & - & 90.9 \\
DGCNN (Wang et al., 2019) & - & - & 92.2 \\
PIDS (Second-order) (Zhang et al., 2023) & 1.25 & - & 92.6 \\
NAS \\
LC-NAS (Li et al., 2022) & 3.61 & & 91.98 \\
PIDS (NAS) (Zhang et al., 2023) & 0.61* & 126 (104+22)* & 92.1* \\
PIDS (NAS, 2x) (Zhang et al., 2023) & 1.31* & 90 (68+22)* & 92.9* \\
\textbf{SHSADE-PIDS (NAS, 1.25X)} & \textbf{0.44}& \textbf{93 (75+18)} & \textbf{92.3} \\
\textbf{SHSADE-PIDS (NAS, 2X)} & \textbf{1.31} & \textbf{93 (70+23)} & \textbf{93.4} \\

\hline
\end{tabular}
\end{table*}


\subsection{Proposed SHSADE-PIDS}
SHSADE-PIDS is an evolutionary algorithm that optimizes neural network architectures by bridging discrete and continuous search spaces. It combines the continuous optimization capabilities of adaptive DE with discrete-continuous mapping to enable efficient architecture search. The overall pipeline is shown in Fig. \ref{fig:shade-pids}.

\begin{enumerate}[(a)] 
\item \textbf{Discrete Configuration Space:} The architecture search space is defined as a set of discrete choices for each architectural parameter, such as a number of filters, kernel size, strides, etc. Each choice is discretized into a predefined set of allowed values $\mathcal{D} = {D_1, D_2, ..., D_m}$ where $D_i = {d_{i1}, d_{i2}, ..., d_{in_i}}$ is the set of $n_i$ possible values for choice $i$.

\item \textbf{Population Initialization:} A population $P$ of $NP$ individuals is randomly initialized, where each individual represents a neural network architecture $x_i$:
\begin{equation}
P = {x_1, x_2, ..., x_{NP}}
\end{equation}
The architecture $x_i$ is a vector of choices $[C_1, C_2, ..., C_m]$ with each $C_i \in D_i$ selected randomly from the allowed discrete values.
\item \textbf{Discrete-Continuous Mapping:} To enable continuous optimization, a mapping $M$ encodes the discrete architecture $x_i$ into a continuous vector $u_i \in [0,1]^m$:
\begin{equation}
u_i = M(x_i)
\end{equation}
For a choice $C_i$ with actual discrete value $k \in D_i$, the mapping is:
\begin{equation}
M(C_i=k) = \frac{\text{Index of } k \text{ in } D_i}{n_i-1}
\end{equation}
This normalization maps the discrete choices to a continuous range while maintaining relative positioning. Gaussian noise is added to $u_i$ for exploration.

\item \textbf{Evaluation:} The population is evaluated using a fast predictive model that estimates the performance of each architecture without full training. This enables quick fitness approximation.

\item \textbf{Evolutionary Optimization} In each generation $g$, SHSADE optimizes the continuous vectors ${u_1, u_2, ..., u_{NP}}$:
\begin{enumerate}[(i)]
\item Select current best architecture $\hat{x}$ as the target vector $\hat{u}$
\item Mutate random subset of $P$ to produce mutated vectors $v_i = \text{SHSADE-Mutation}(u_i)$
\item Crossover mutated vectors with target: $z_i = \text{DE-Crossover}(v_i, \hat{u})$
\item Decode trial vectors into architectures: $\hat{x}_i = M^{-1}(z_i)$
\item Evaluate trial architectures and update $P$
\end{enumerate}

The mutation strategy adapts over generations based on successful parameter changes stored in an external archive. The evolution gradually improves the population, converging towards optimal architectures.

\item \textbf{Termination and Result Extraction:} After a termination criterion is met, the overall best architecture $\hat{x}^*$ is returned as the final result of the search
\end{enumerate}


In summary, SHSADE-PIDS bridges discrete and continuous spaces through mapping, leveraging the exploration capabilities of adaptive DE to efficiently search neural architectures. The combination of discrete configuration, continuous-discrete mapping, fast evaluation, and adaptive evolution enables effective architecture optimization.




\begin{figure}[h]
\centering
\includegraphics[width=\linewidth]{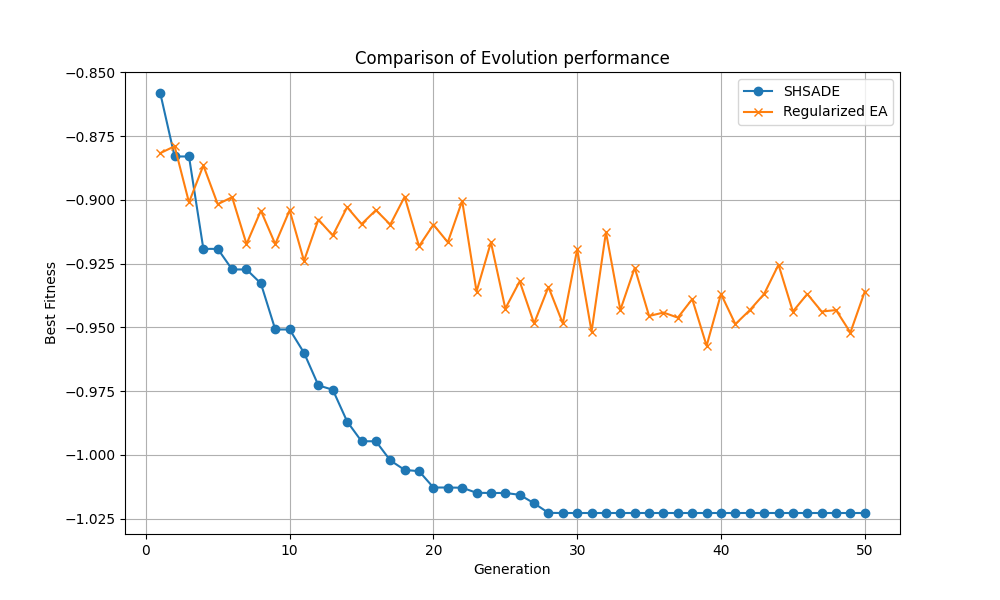}
\caption{The evolutionary process of the proposed SHSADE and the regularized EA used in PIDS (NAS)~\cite{zhang2023pids}.}
\label{fig:evoCurve}
\end{figure}

\begin{figure}[h]
\centering
\includegraphics[width=0.9\linewidth]{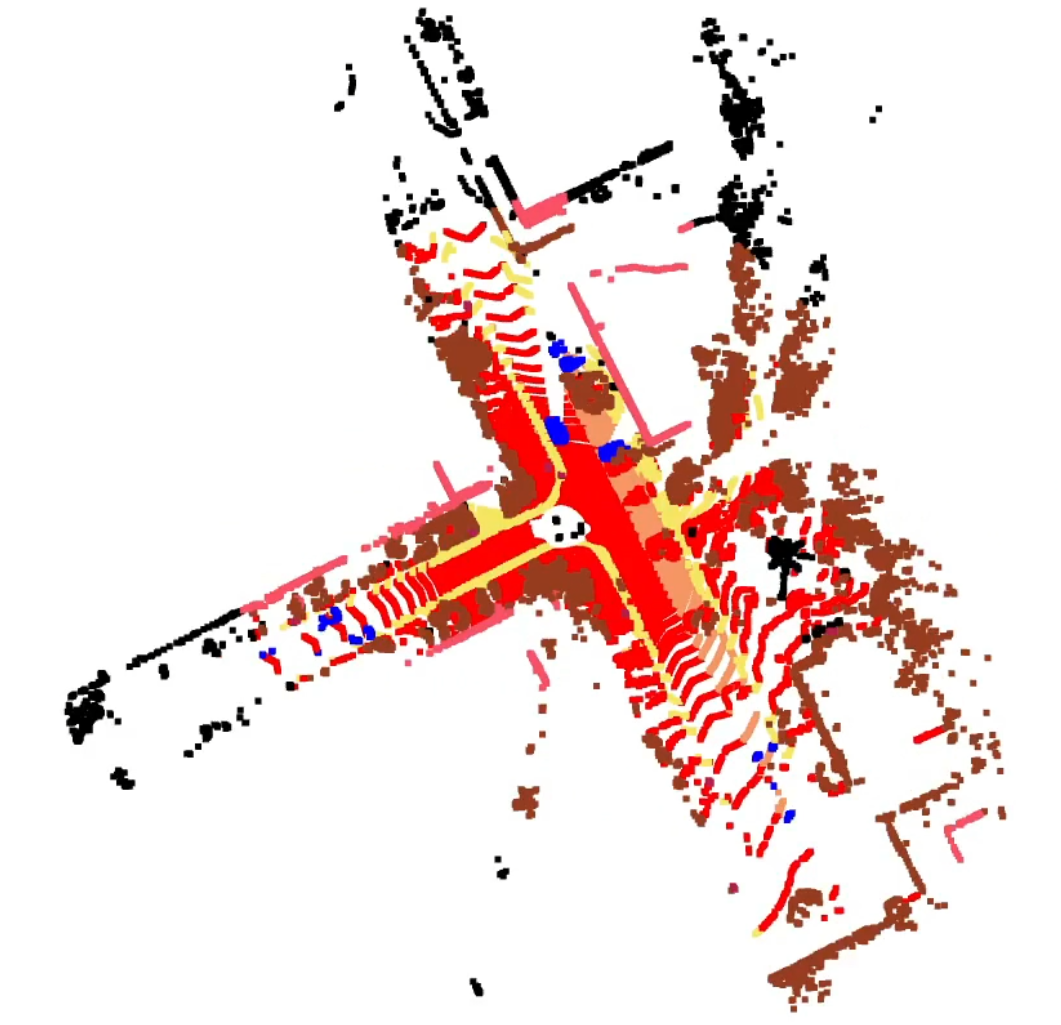}
\caption{One frame from Semantic segmentation result of sequence 08 in SemanticKITTI.}
\label{fig:semantic_segmentation}
\end{figure}



\section{Experiment}
\textbf{Dataset:} We evaluate our method on two popular 3D deep learning benchmarks: ModelNet40 for 3D object classification and SemanticKITTI for semantic segmentation of LiDAR scans. ModelNet40 contains 12,311 CAD models from 40 categories. We follow the official split of 9,843 models for training and 2,468 for testing. SemanticKITTI has over 43,000 densely annotated LiDAR scans covering a total of 39.2 km across various urban and highway environments. Each scan has ~130,000 points labeled into 28 semantic classes.

\textbf{Implementation Details:} Our SHSADE-guided neural architecture search is implemented in PyTorch. All experiments are conducted on a server with an NVIDIA RTX 4090 GPU. For a fair comparison, we keep the training configurations the same between our method and baseline NAS strategies. Specifically, we use a mini-train/mini-val split of 80/20 and train for 300 epochs using an Adam optimizer with an initial learning rate of 0.001 and weight decay of 0.0001. During the search, 100 independent SHSADE runs are performed to discover top architectures. The search budget is limited to 500 sampled architectures.0

Fig. \ref{fig:evoCurve} shows the comparison of evolution performance for the SHSADE algorithm and the baseline work, the regularized EA used in PIDS (NAS)~\cite{zhang2023pids}. Obviously, SHSADE achieved a better convergence.

\section{Result}

\subsection{Semantic Segmentation}
Table \ref{table:semantic_kitti} summarizes the mIOU on sequence 08 (validation split) of SemanticKITTI. For segmentation, SHSADE-PIDS attains a leading mIoU of 64.51\% on the challenging SemanticKITTI benchmark, surpassing prior published methods. Critically, this is achieved using a highly compact model, with only 0.55M/1.36M parameters. This represents a 22-26X parameter reduction compared to other top-performing approaches like KPConv. The small model size enables more efficient inference and deployment. SHSADE-PIDS further reduces computational overhead, requiring just 4.5G/8.6G MACs versus heavier models like SalsaNext and MinkowskiNet, which demand 6-14X more operations. Latency is also competitive at 119ms/132ms, on par or faster than most methods.

\subsection{Classification}
Table \ref{table:modelnet40} presents the overall accuracy of ModelNet40. For classification, SHSADE-PIDS again demonstrates optimized efficiency-accuracy trade-offs. With only 0.44M/1.31M parameters, it achieves top accuracy of 92.3\%/93.4\% on ModelNet40, while having up to 34X fewer parameters than KPConv. Latency remains low at 93ms for both model sizes.

The results demonstrate that the proposed SHSADE-PIDS method achieves better performance for 3D point cloud segmentation and classification while optimizing model efficiency.

\section{Conclusion}
This paper has presented SHSADE-PIDS, an evolutionary neural architecture search method that bridges discrete and continuous spaces to optimize neural networks for 3D point cloud tasks. SHSADE-PIDS combines the continuous optimization strengths of SHSADE with a discrete-continuous mapping to enable efficient exploration of architecture configurations.
Comprehensive experiments on semantic segmentation using SemanticKITTI and shape classification using ModelNet40 demonstrate SHSADE-PIDS's capabilities in discovering specialized models for point cloud processing. The neural architectures found by SHSADE-PIDS significantly advance higher accuracy and efficiency over prior works, including both hand-designed networks and other NAS approaches.
Specifically, SHSADE-PIDS attains leading segmentation performance with just 0.55M parameters and 4.5GMACs, reducing overhead by over 22-26X versus other top methods while achieving higher 64.51\% mIoU. For classification, it secures a higher accuracy of 93.4\% with only 1.31M parameters, surpassing larger models.
In conclusion, this work provides a new perspective on effectively leveraging continuous EA variants for discrete architecture search. The proposed SHSADE-PIDS approach and analyses offer valuable insights into hybridizing evolutionary algorithms with neural architecture optimization, particularly for emerging problem domains like point cloud processing.

\bibliographystyle{IEEEtran}
\bibliography{sn-bibliography}

\end{document}